\documentclass[pdflatex,sn-mathphys-num]{sn-jnl}

\usepackage{graphicx}%
\usepackage{multirow}%
\usepackage{amsmath,amssymb,amsfonts}%
\usepackage{amsthm}%
\usepackage{mathrsfs}%
\usepackage[title]{appendix}%
\usepackage{xcolor}%
\usepackage{textcomp}%
\usepackage{manyfoot}%
\usepackage{booktabs}%
\usepackage{algorithm}%
\usepackage{algorithmicx}%
\usepackage{algpseudocode}%
\usepackage{listings}%
\usepackage{makecell}
\usepackage{siunitx}

\usepackage{chngcntr}

\renewcommand{\figurename}{Figure}

\begin{document}


\title[SAM]{SAMJ: Fast Image Annotation on ImageJ/Fiji via Segment Anything Model}


\author[1,2,3]{\sur{Carlos García-López-de-Haro}}

\author[1,2]{\sur{Caterina Fuster-Barceló}}

\author[4]{\sur{Curtis T. Rueden}}

\author[5]{\sur{Jónathan Heras}}

\author[6,7]{\sur{Vladim\'ir Ulman}}

\author[8,9,10]{\sur{Daniel Franco-Barranco}}

\author[5]{\sur{Adrián Inés}}

\author[4]{\sur{Kevin W. Eliceiri}}

\author[1, 3, 11]{\sur{Jean-Christophe Olivo-Marin}}

\author[12]{\sur{Jean-Yves Tinevez}}

\author*[13]{\sur{Daniel Sage}}

\author*[1,2]{\sur{Arrate Muñoz-Barrutia}}

\affil[1]{\orgdiv{Bioengineering Department}, \orgname{Universidad Carlos III de Madrid}, \orgaddress{\city{Leganes}, \country{Spain}}}

\affil[2]{\orgname{Instituto de Investigación Sanitaria Gregorio Marañón}, \orgaddress{\city{Madrid}, \country{Spain}}}

\affil[3]{\orgdiv{Bioimage Analysis Unit}, \orgname{Institut Pasteur, Université Paris Cité}, \orgaddress{\city{Paris}, \country{France}}}

\affil[4]{\orgdiv{Center for Quantitative Cell Imaging}, \orgname{University of Wisconsin}, \orgaddress{\city{Madison}, \country{USA}}}

\affil[5]{\orgdiv{Department of Mathematics and Computer Science}, \orgname{University of La Rioja}, \orgaddress{\city{Logroño}, \country{Spain}}}

\affil[6]{\orgdiv{IT4Innovations}, \orgname{VSB -- Technical University of Ostrava}, \orgaddress{\city{Ostrava}, \country{Czech Republic}}}

\affil[7]{\orgdiv{Central European Institute of Technology (CEITEC)}, \orgname{Masaryk University}, \orgaddress{\city{Brno}, \country{Czech Republic}}}

\affil[8]{\orgname{MRC Laboratory of Molecular Biology}, \orgaddress{\city{Cambridge}, \country{UK}}}

\affil[9]{\orgdiv{Department of Physiology, Development and Neuroscience}, \orgname{University of Cambridge}, \orgaddress{\city{Cambridge}, \country{UK}}}

\affil[10]{\orgname{Donostia International Physics Center (DIPC)}, \orgaddress{\city{San Sebastian}, \country{Spain}}}

\affil[11]{\orgdiv{CNRS UMR 3691}, \orgname{Institut Pasteur}, \orgaddress{\city{Paris}, \country{France}}}

\affil[12]{\orgdiv{Image Analysis Hub}, \orgname{Institut Pasteur, Université Paris Cité}, \orgaddress{\city{Paris}, \country{France}}}

\affil[13]{\orgdiv{Biomedical Imaging Group and Center for Imaging}, \orgname{Ecole Polytechnique Fédérale de Lausanne (EPFL)}, \orgaddress{\city{Lausanne}, \country{Switzerland}}}


\abstract{

Mask annotation remains a significant bottleneck in AI-driven biomedical image analysis due to its labor-intensive nature. To address this challenge, we introduce SAMJ, a user-friendly ImageJ/Fiji plugin leveraging the Segment Anything Model (SAM). SAMJ enables seamless, interactive annotations with one-click installation on standard computers. Designed for real-time object delineation in large scientific images, SAMJ is an easy-to-use solution that simplifies and accelerates the creation of labeled image datasets.
}



\maketitle


Access to high-quality annotated images is key to training and developing new methods for biomedical image analysis and object segmentation. However, creating large datasets of annotated images is a time-consuming, labor-intensive, and subjective process~\cite{pelt2020tackling}. The demand becomes even more pronounced when training supervised Deep Learning (DL) models, typically requiring hundreds to thousands of annotated images~\cite{zulueta2023mifa}. Therefore, there is a critical need for efficient annotation tools capable of running on commonly available computers. Such tools must reduce IT-related barriers, ensuring broader accessibility for domain-specific experts such as biologists. 

These solutions should also be compatible with widely used bioimage software platforms, such as ImageJ/Fiji~\cite{schindelin2012fiji}, and must be fast enough to accelerate the annotation of dozens of objects in large images, which are typical in biological research and essential for training machine learning models.

Recently, Meta Research released the Segment Anything Model (SAM)~\cite{kirillov2023segment}, a foundational DL framework designed for one-shot and promptable image segmentation. SAM represents a significant breakthrough, capable of 
delineating complex objects and enabling semi-automated, interactive annotation across diverse image types. In its standard mode of operation, SAM requires user-defined prompts, such as points, sets of points, or bounding boxes, to efficiently isolate target objects.

SAM has been made available in commonly used bioimage software packages, such as QuPath~\cite{qupath_sam_plugin} and Napari~\cite{archit2025segment}, providing interactive annotation, but has not yet been incorporated into the ImageJ ecosystem~\cite{imagej}. Other proprietary software, such as Matlab, ArcGIS, Kili Technology, or Unitlab Magic Touch have incorporated SAM as an add-on tool to enhance their segmentation capabilities. Similarly, Segment Anything Annotator\footnote{\url{https://github.com/haochenheheda/segment-anything-annotator}}, an open-source Python-based UI application, uses SAM for pixel-level annotation.

Although many annotation tools are both user-friendly and widely accessible, they often face challenges when dealing with uneven illumination, overlapping structures, or subtle intensity gradients. Several plugins have introduced more advanced methods: AnnotatorJ~\cite{annotatorj}, a popular Fiji plugin,  supports semi-automatic cellular annotation by combining manual input with deep learning; Labkit~\cite{labkit} leverages machine learning for pixel-wise classification; and Thomas et al.~\cite{thomas2021fiji} propose a streamlined approach to systematic manual annotation by assigning predefined categories to images or regions. However, none of these plugins yet benefit from the advantages that SAM provides.

\begin{figure}[h]
    \centering
    \includegraphics[width=1\linewidth]{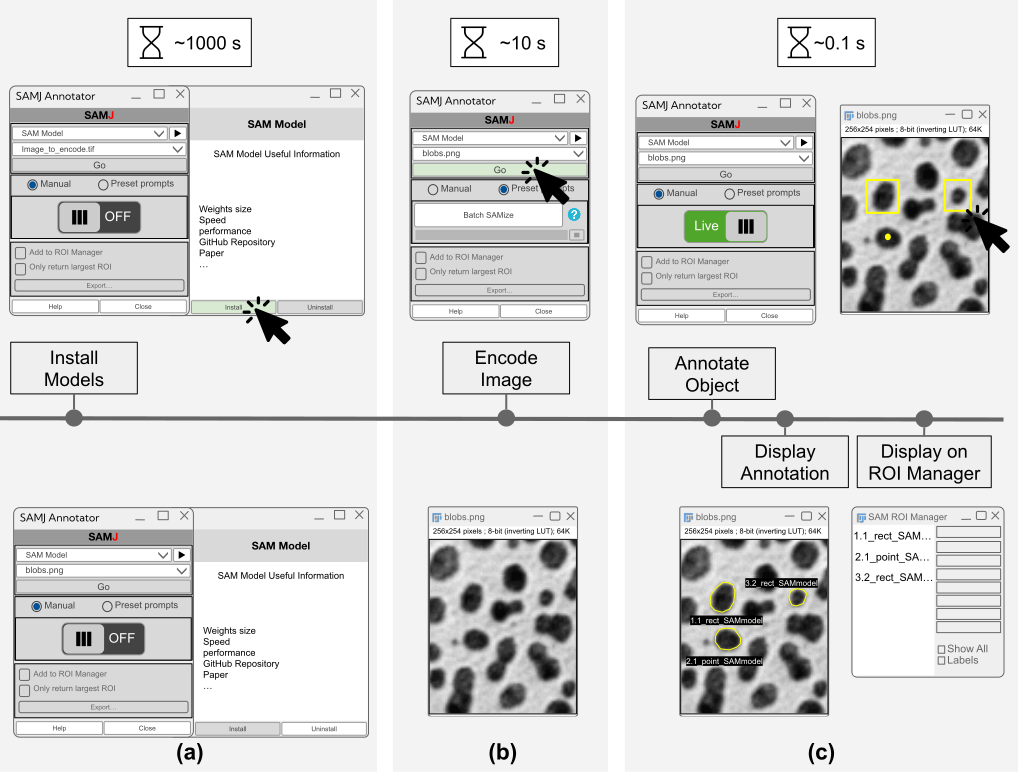}
    \caption{\textbf{SAMJ User Interface and Workflow with estimated time.} This figure illustrates the typical workflow of SAMJ for annotating objects in an image, including the time required for each step on standard workstations. (a) Model Installation: The most time-consuming step, where SAMJ installs the selected model and, if necessary, sets up the environment, taking approximately 1000 seconds. (b) Image Encoding: When the user clicks “Go” in the plugin, the image is processed to generate an embedding, which takes around 10 seconds. (c) Object Annotation: Since no re-embedding is required, this step is immediate, with each annotation generated in approximately 0.1 seconds per user click. The annotation process can be repeated multiple times as desired for different objects, offering rapid and interactive segmentation.}
    \label{fig:samj-details}
\end{figure}

In this manuscript, we present SAMJ, a plugin that brings the capabilities of SAM into the Fiji ecosystem, both as a standalone  Fiji plugin and as an extension for Labkit\cite{labkit}.
We designed SAMJ for one-click installation. It features a no-code interface, and runs on standard computers without requiring specialized hardware. By lowering technical barriers, it broadens access to cutting-edge AI tools within the life sciences community. A key contribution of our work is the tailored adaptation of SAMJ to efficiently handle scientific images. By harnessing the powerful segmentation capabilities of SAM, this tool has the potential to significantly accelerate the annotation process, enabling users to instantly generate regions of interest around target objects with a single point click or bounding box.


SAMJ leverages the core architecture of SAM, which is built upon three key components: an image encoder, a prompt encoder, and a mask decoder. The image encoder is based on a Visual Transformer (ViT) \cite{dosovitskiy2021an} that encodes the input image, offering high representational power at the expense of significant computational resources. The prompt encoder is another transformer to encode the prompt given by the user to point or ``hint'' to the location of the object of interest. Finally, the encoded image and prompt are used by the mask decoder, which combines elements of both a transformer and a Convolutional Neural Network, to generate the mask of interest.

The SAMJ plugin is installed as any standard Fiji extension, eliminating the technical IT barriers often associated with Python packages and making advanced image segmentation accessible to a wider audience. Unlike most Python-based libraries that require command-line interfaces for environment configuration and dependency management, SAMJ simplifies the process with a seamless, one-click installation, enabling effortless deployment. 

In addition, five SAM variants are available through SAMJ. Each of these models can run on standard workstations or mid-range laptops, eliminating the need for high-end computational resources that the original SAM typically requires. These variants are SAM-2~\cite{ravi2024sam}, EfficientSAM~\cite{xiong2024efficientsam}, and EfficientViTSAM~\cite{zhang2024efficientvit}.

The annotation process in the SAMJ plugin mirrors SAM’s workflow (\autoref{fig:samj-details}). Each image is encoded once by the image encoder, the most computationally intensive step. Afterwards, prompts are processed and masks generated almost instantly, allowing users to interactively produce multiple annotations with real-time responsiveness.
SAMJ integrates seamlessly with Fiji’s existing tools, allowing users to create prompts such as points and rectangles directly through Fiji's familiar toolbar, thereby minimizing the learning curve for new users.

The SAMJ plugin offers two annotation modes: Live and Batch mode (called BatchSAMize). In Live mode, prompts are drawn directly on the encoded image, and annotations are generated interactively, one at a time. In Batch mode, users can define multiple seed points to serve as prompts for segmenting multiple objects in a raw image. These seeds can be generated automatically using traditional Fiji commands (e.g., Find Maxima, Watershed) within an ImageJ macro, followed by a call to SAMJ’s Batch mode to segment the whole image.
Additionally, users can enhance annotations by providing seeds from other segmentation methods as prompts, enabling iterative refinement.

Beyond its ease of use, SAMJ is designed for adaptability and seamless integration. Powered by SAM, it can enhance annotation workflows across a range of Java-based platforms, for example, in addition to Labkit, we have also integrated it into BigDataViewer. \cite{pietzsch2015bigdataviewer}. Moreover, to support broader adoption, SAMJ provides a well-documented API, enabling developers to easily incorporate SAM functionality into their preferred Java software effortlessly. Furthermore, it includes a software-agnostic Java GUI, facilitating straightforward integration into any Java-based environment.

\begin{figure}
    \centering
    \includegraphics[width=1\linewidth]{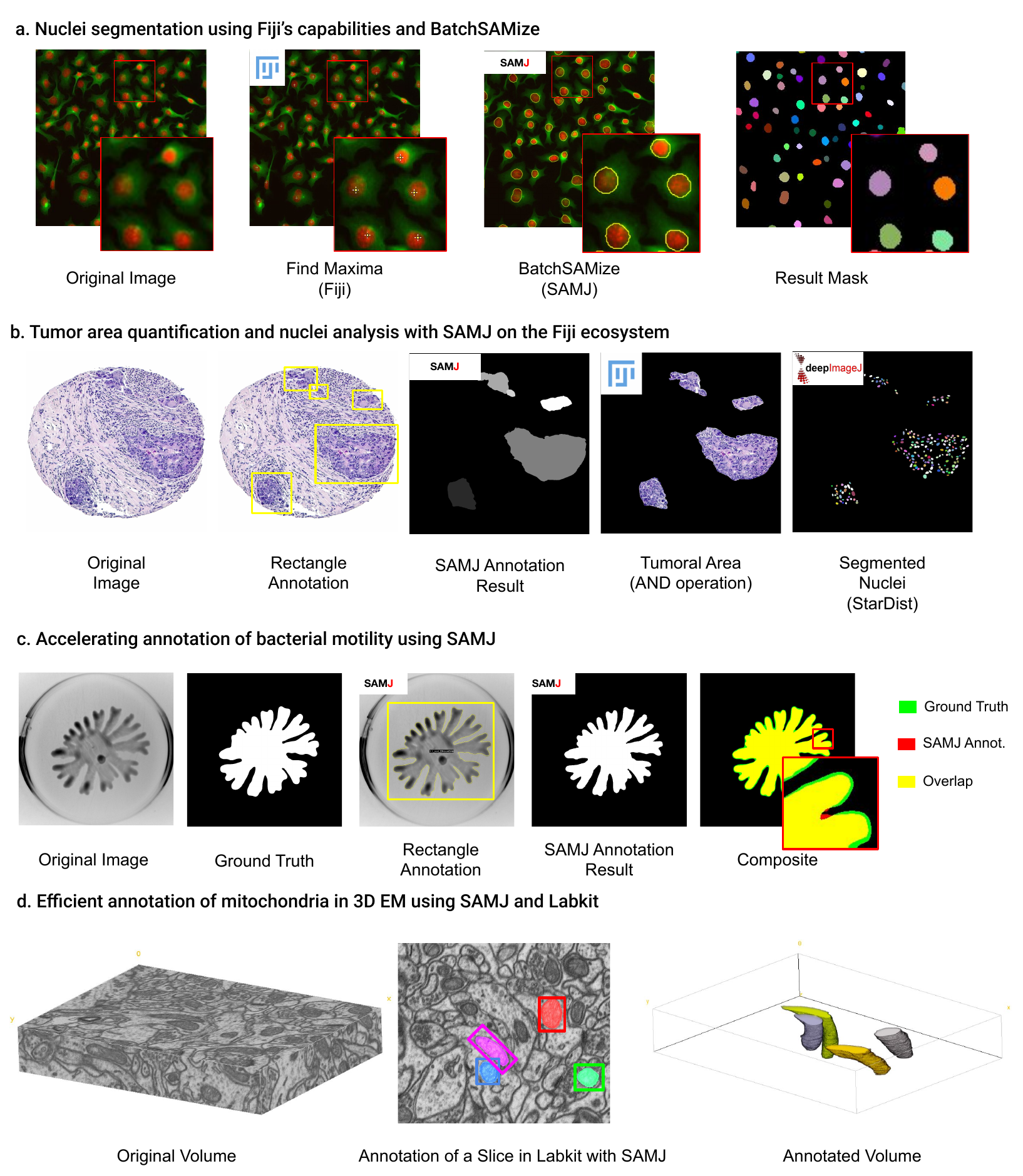}
    \caption{\textbf{Use cases demonstrating the versatility of SAMJ for bioimage analysis.} (a) Nuclei Segmentation using Fiji's capabilities and BatchSAMize: An image from the CellPose dataset~\cite{stringer2021cellpose} is used to demonstrate the segmentation of nuclei in the red channel. Pixel intensity maxima are identified in Fiji to generate single-point prompts for each nucleus. These prompts are processed in batch mode using SAMJ, resulting in semantic segmentation of individual nuclei. (b) Tumor Area Quantification and Nuclei Analysis with SAMJ and StarDist plugins for Fiji: Breast cancer TMA images, stained with H\&E provided by the British Columbia Cancer Agency (BCCA)~\cite{shamai2022deep}, are used to quantify tumoral regions and their nuclei. Tumoral areas are annotated with SAMJ's rectangle prompt, generating masks that are combined with the original image in Fiji using an AND operation to obtain the intersection. StarDist is then applied through deepImageJ to segment individual nuclei in the tumoral areas. (c) Accelerating annotation of Bacterial Motility: SAMJ is applied to annotate motile bacteria using a single rectangle annotation. When compared to Ground Truth, SAMJ’s annotations achieve comparable or superior precision, showcasing its efficiency in handling complex shapes and streamlining high-throughput workflows. Each use case highlights SAMJ’s integration with Fiji, combining SAM’s advanced annotation capabilities with Fiji’s extensive image processing tools. (d) Efficient annotation of 3D Electron Microscopy images of mitochondria~\cite{casser2020fast} using SAMJ and Labkit: SAMJ is integrated into Labkit to support 3D and multi-label annotation. Users can annotate structures on arbitrarily oriented and scaled slices, improving visibility and accuracy of objects with complex spatial orientation. The multi-class labeling capability of Labkit allows the annotation of several distinct structures within the same volume. This workflow reduces annotation effort and ensures spatial consistency across slices.
    }
    \label{fig:panel-1}
\end{figure}

A key strength of SAMJ is its integration of Python methods into Java environments. This is achieved through Appose\footnote{\url{https://github.com/apposed/appose}}, a Java package that allows Java and Python to run as separate processes yet communicate in real time. Leveraging Micromamba\footnote{\url{https://mamba.readthedocs.io/en/latest/user_guide/micromamba.html}}, Appose also automates Python environment setup, totally eliminating manual configuration via the command line. As a result, SAMJ is remarkably user-friendly, granting effortless access to advanced methods.

To illustrate the power of SAMJ and its seamless integration with the Fiji ecosystem, we present four representative use cases highlighting its capabilities (\autoref{fig:panel-1}). These examples illustrate SAMJ's ability to adapt to diverse image annotation tasks while taking advantage of Fiji's powerful image processing tools. In all cases, the targets are compact and well-defined--conditions under which SAMJ performs particularly well. Elongated or branched structures may pose more of a challenge; However, users can refine prompts, adjust scale settings, or combine SAMJ with complementary methods. Experimentation is encouraged, as SAMJ often proves effective even under suboptimal imaging conditions.

In the first use case, we illustrate nuclei segmentation in a fluorescence image using the BatchSAMize mode of SAMJ. Classical Fiji commands are used to detect nuclei, which serve as single-point prompts for batch processing in SAMJ. This workflow highlights the synergy between the preprocessing capabilities of Fiji and the segmentation power of SAMJ, streamlining large-scale annotation tasks. 

The second use case intended for tumor area quantification focuses on breast cancer TMA images stained with H\&E. The task involved quantifying tumoral areas and the number of nuclei within these regions. Using SAMJ's rectangle annotation feature, tumoral regions were outlined, generating masks for these areas. Subsequently, individual nuclei were segmented with StarDist~\cite{schmidt2018cell} using the deepImageJ~\cite{gomez2021deepimagej}~\cite{imagej} plugin. This workflow demonstrates SAMJ’s role as a flexible and efficient annotation tool that seamlessly integrates with Fiji, enabling efficient annotation and precise quantification for tumor analysis.

The third use case highlights SAMJ's efficiency in annotating complex shapes~\cite{motilityj}. Bacteria from motility studies were annotated using a single rectangle prompt per image, significantly reducing manual effort. When compared to manually generated ground truth, SAMJ’s annotations are visually indistinguishable. This use case demonstrates how SAMJ accelerates annotation tasks for intricate biological structures, making it a powerful tool for high-throughput studies.

The final use case demonstrates SAMJ's capability for efficient annotation of mitochondria instances in 3D electron microscopy volumes through its integration with Labkit. Annotating objects in 3D is inherently challenging due to the difficulty of maintaining spatial coherence across slices and the substantial effort required for manual annotation. SAMJ streamlines this process, and its integration into Labkit--designed for interactive 3D visualization--enables users to annotate structures on arbitrarily oriented slices, ensuring optimal views of the object of interest. Moreover, Labkit support for multi-class labeling allows users to assign distinct labels to multiple structures within the same volume. This combination significantly reduces annotation effort while improving both accuracy and consistency, highlighting the power of integrating SAMJ with complementary Fiji tools for complex 3D bioimage annotation tasks.

Together, these use cases highlight the versatility and strength of SAMJ integration within the Fiji ecosystem. By combining the advanced segmentation capabilities of SAM with the rich set of image processing tools of Fiji, SAMJ enables more efficient, precise, and scalable annotation workflows across a wide range of biological imaging tasks. This integration lowers technical barriers and supports diverse use cases, from 2D fluorescence images to complex 3D volumes. Consequently, it empowers the broader life science community--particularly biologists who primarily rely on GUI software platforms--with powerful deep-learning tools that would otherwise remain accessible only to a few accelerating and improving their bioimage analysis.


Accurate annotation is fundamental to deep learning-based image analysis, playing a critical role in model training, validation, and fine-tuning. To address the limitations posed by the high embedding latency of SAM, particularly in large microscopy images, we developed SAMJ, an intuitive and interactive annotation tool tailored for the life sciences. SAMJ integrates efficient SAM2 models optimized to run on mid-range computers and includes a Java-Python bridge to eliminate complex installation steps. As an ImageJ/Fiji plugin, it remains fully compatible with standard Fiji tools such as ROI selection, the ROI manager, and macros. Through this seamless integration, SAMJ enables efficient, high-quality annotations and supports the creation of curated datasets, thus contributing to the development of more accurate and robust models. 

In conclusion, SAMJ provides a user-friendly and accessible way for biologists and bioimage analysts to adopt cutting-edge AI methods, facilitating faster and more precise image annotation and ultimately accelerating scientific discovery.

\bmhead{Acknowledgements}

Views and opinions expressed are however those of the author(s) only and do not necessarily reflect those of the European Union or the European Research Council Executive Agency. Neither the European Union nor the granting authority can be held responsible for them.

\bmhead{Author contributions}

Conceptualization: D.S., C.G-L-H. and A.M-B. Software: C.G-L-H., V.U., D.S and C.T.R. Interface design: D.S., C.G-L-H., C.T.R. and C.F-B. Validation and testing: C.G-L-H., C.F-B., V.U., D.S and D.F-B.  Writing: C.G-L-H., C.F-B., V.U., A.M-B. and D.S. Writing—review and editing: all authors. Supervision: A.M-B. and D.S.

\bmhead{Funding statement}

This work was partially funded by the project CNRS France 2030 Recherche à Risque (RI2) ProteoVir (J-C.O-M. and C.G-L-H).
This work was partially supported by the European Union’s Horizon Europe research and program under grant agreement number 101057970 (AI4Life project) awarded to A.M-B. and the Ministerio de Ciencia, Innovación y Universidades, Agencia Estatal de Investigación, MCIN/AEI/10.13039/501100011033/, under grants PID2019-109820RB and PID2023-152631OB-I00, co-financed by European Regional Development Fund (ERDF), “A way of making Europe”.
This work has been funded in part by the Agence Nationale de la Recherche through the programs PR[AI]RIE-PSAI (ANR-23-IACL-0008) and France-BioImaging infrastructure (ANR-10-INBS-04) (J-C.O-M. and J-Y.T.). This project has received funding from the Innovative Medicines Initiative 2 Joint Undertaking under grant agreement No 945358 (BIGPICTURE). This Joint Undertaking receives support from the European Union’s Horizon 2020 research and innovation program and EFPIA (J-C.O-M.).
J-C.O-M and J-Y.T were also supported by the French National Research Agency (France BioImaging, ANR-24-INBS-0005 FBI BIOGEN).
V.U. was supported by grant number 2024-342803 from the Chan Zuckerberg Initiative DAF, an advised fund of Silicon Valley Community Foundation.
C.T.R. and K.W.E were funded by the National Institutes of Health, grant number NIH P41 GM135019.
J. H. and A. I. were supported by Agencia de Desarrollo Económico de La Rioja ADER 2022-I-IDI-00015 and by project AFIANZA 2024/01 granted by the Autonomous Community of La Rioja.
J. H. was also supported by the Government of La Rioja through Proyecto Inicia 2023/01.

\begin{appendices}

\counterwithin{figure}{section} 
\counterwithin{table}{section}

\section{Online Methods}\label{secA1}
\subsection{Review of Annotation Tools for Image Segmentation}

We provide a comprehensive overview of the existing annotation tools for bioimage analysis, summarizing key features, such as their release dates, supported annotation types, and integration with the Segment Anything Model (SAM) or any other AI assistant (Table~\autoref{tab:literature-review}). This table highlights the diversity of tools available across platforms like Fiji, QuPath, and Napari, as well as their varying levels of SAM integration. Notably, while tools such as QuPath and Napari have embraced SAM, Fiji-based plugins like AnnotatorJ lack direct SAM support. SAMJ addresses this gap by bringing SAM’s advanced annotation capabilities to the Fiji environment, combining flexibility with ease of use. This comparative analysis underscores SAMJ’s unique contribution to enabling SAM-powered annotation within the Fiji ecosystem.

\begin{table}
    \centering
    \begin{tabular}{|c|c|c|c|c|c|c|}
    \toprule
        \makecell{\textbf{Tool/}\\\textbf{Plugin}} & \textbf{Platform} & \textbf{Release} & \makecell{\textbf{Last}\\\textbf{Update}} & \textbf{AI Assistance} & \textbf{Annotation} & \makecell{\textbf{Programming}\\\textbf{Language}} \\
    \midrule
        MicroSAM & Napari & 2023 & 03/2025 & \makecell{Yes\\\footnotesize(SAM fine-tune)}  & \makecell{Polygons\\ BBox \\ Points} & Python \\ \hline
        
        \makecell{QuPath\\Ext. SAM} & QuPath & 2023 & 09/2024 & \makecell{Yes\\\footnotesize(SAM2)} & \makecell{AutoMask\\BBox\\Poitns} & Python\\ \hline
        
        \makecell{Napari Plugin\\of SAM} & Napari & 2023 & 04/2023 & \makecell{Yes\\\footnotesize(SAM1)} & \makecell{BBox\\Points} & Python \\ \hline
        
        \makecell{Napari Plugin\\of SAM2} & Napari & 2024 & 09/2024 & \makecell{Yes\\\footnotesize(SAM2)} & \makecell{BBox\\Points} & Python \\ \hline
        
        AnnotatorJ & Fiji & 2020 & 10/2020 & \makecell{Yes\\(CNN)} & \makecell{Instance\\Semantic\\BBox} & Java \\ \hline
        
        Labkit & Fiji & 2022 & 10/2024 & \makecell{Yes\\(Random Forest\\SAM1,2 )} & \makecell{Pixel\\Classification} & Java \\\hline
        
        \makecell{Qualitative\\Annotations} & Fiji & 2020 & 04/2021 & \makecell{Yes\\(CNN)} & \makecell{One or\\Multiple ROI} & Python\\ \hline
         
        SAMJ & Fiji & 2025 & 04/2025 & \makecell{Yes\\\footnotesize(SAM1,2)} & \makecell{BBox\\Points} & Java\\
    \bottomrule
    \end{tabular}
    \caption{\textbf{Summary of tools integrating annotation capabilities in bioimage analysis platforms.} The table highlights the main tools and plugins used for annotation in various bioimage analysis platforms, detailing their release (YYYY) and last update dates (MM/YYYY, Last accessed: 20MAY2025), annotation features, integration with the SAM, and the programming languages used for their development. SAM integration is specified with its supported annotation types, including bounding boxes (BBox), points, polygons, and more, showcasing the variety of annotation modes available.}
    \label{tab:literature-review}
\end{table}

\subsection{Available SAM Variants}

The original SAM model is computationally intensive and was initially introduced in three versions, differentiated by the size of the Vision Transformer (ViT): base, large, and huge. However, even the smallest model, SAM-base, is too heavy to run on standard CPUs due to its high resource demands.

To address these limitations, newer and more lightweight variants have been developed. These models (efficientSAM \cite{xiong2024efficientsam}; efficientViTSAM \cite{zhang2024efficientvit}) achieve a significant improvement in computational efficiency with only a minimal trade-off in annotation quality, making SAM-like models accessible for lower-end CPUs.

Moreover, the release of the second generation of the original SAM, known as SAM-2 \cite{ravi2024sam}, marked a breakthrough. This new family of models not only enhances annotation quality but also significantly reduces computational costs.

For these reasons, the list of models available in SAMJ does not include any of the original SAM versions. This list is expected to evolve as new models achieve improvements in performance and efficiency. Currently, the supported models in SAMJ include: \textbf{SAM-2 (tiny, small, and large)}, \textbf{EfficientSAM}, and \textbf{EfficientViTSAM}. For the most up-to-date list, please refer to the official SAMJ GitHub repository: \url{https://github.com/segment-anything-models-java/SAMJ/blob/main/README.md}.

\begin{table}[htbp]
    \centering
    \caption{Encoding time comparison for SAM-2, EfficientSAM, and original SAM models. 
The table shows model sizes (in MB) and the time required (in seconds) to perform encoding. 
Lightweight models such as SAM-2 Tiny and EfficientViTSAM-l2 significantly reduce both size and encoding time compared to the original SAM models. All results in this table were obtained on a workstation equipped with a 13th Gen Intel® Core™ i7-13700H processor and 32 GB of RAM. Although running the model on different hardware may produce varying absolute timings, the relative performance differences are expected to remain consistent.}
    \label{tab:model_times}
    \begin{tabular}{lccc}
        \toprule
        \textbf{} & \textbf{Model Size (MB)} & \textbf{Time to encode (seconds)} & \textbf{In SAMJ} \\
        \midrule
        SAM-2 Tiny & 148.7 & 1.14 & Yes \\
        SAM-2 Small & 175.8 & 1.39 & Yes \\
        SAM-2 Large & 856.4 & 5.59 & Yes \\
        EfficientSAM & 105.7 & 3.19 & Yes \\
        EfficientViTSAM-l2 & 245.7 & 0.74 & Yes \\
        SAM Base & 375.0 & 15 & No \\
        SAM Large & 1191.64 & 15 & No \\
        SAM Huge & 2445.74 & 15 & No \\
        \bottomrule
    \end{tabular}
\end{table}

\subsection{SAMJ Embedding Strategy} 

To generate annotations, SAM must first embed the image, or at least a portion of it. This step creates an embedding of the image, which is then used to generate subsequent annotations. The region for which the embedding has been computed determines the area SAMJ can work with to create annotations. If the user navigates to a region of the image that has not yet been processed, the embedding will need to be recalculated for that specific region before new annotations can be generated.

\begin{figure}
    \centering
    \includegraphics[width=0.9\linewidth]{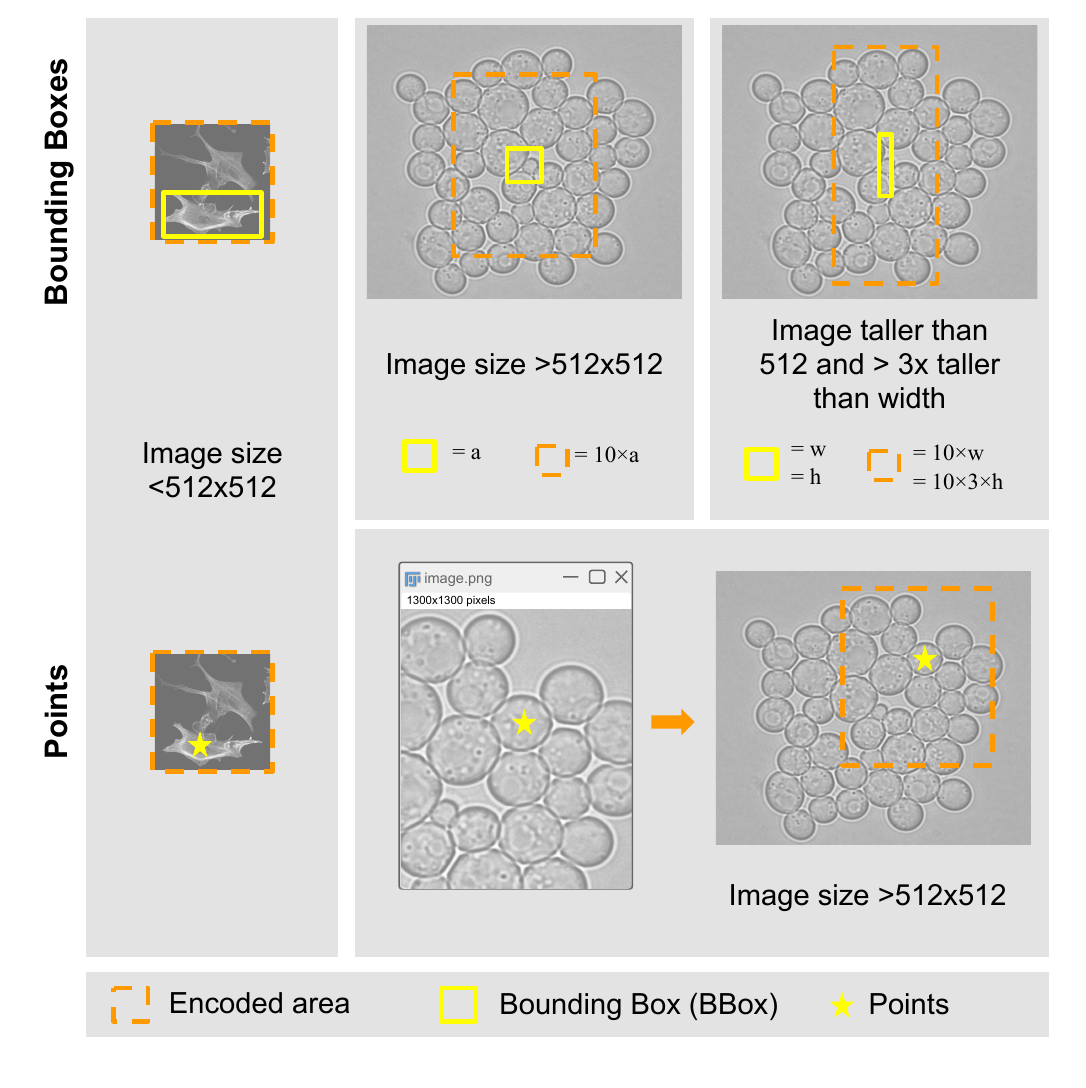}
    \caption{\textbf{Embedding strategies.} SAMJ generates embeddings for regions of the image based on the user-defined input and the size of the image. For smaller images (smaller than $512\times512\ \text{pixels}$), the entire image is embedded regardless of the prompt. For larger images (bigger than $512\times512\ \text{pixels}$), different strategies are followed depending on the prompt and size of the bounding box. If the bounding box is relatively regular, the embedded area will be ten times the size of the bounding box. If one side of the bounding box is three times smaller than the other side of the bounding box, the embedded area will extend ten times the length of the smaller side and ten-thirds (approximately 3.33 times) the length of the larger side. For larger images and point prompts, the embedded area will include the visible region in the Fiji window plus an additional margin.}
    \label{fig:embeddings}
\end{figure}

Having a clear strategy to define the portion of the image for which the embedding is created is therefore critical. In SAMJ, this is achieved by leveraging the interactive features provided by Fiji, such as the region of the image currently being viewed and user input. Depending on the user's input, the method for determining the portion of the image to be embedded varies (see \autoref{fig:embeddings}). This strategy aims to enhance the quality of annotations produced by SAMJ and enables annotations to be created for a wider range of object shapes and sizes.

Consequently, there may be slight differences between annotations generated by SAMJ and those produced using the original SAM implementation. These differences reflect the flexibility of SAMJ’s embedding strategy in accommodating diverse annotation requirements.

\subsection{Bounding Box Size vs. Image Size}

SAM and its variants were primarily trained on natural images rescaled to a 1024×1024 pixel size. Under these training conditions, objects generally maintain a consistent size ratio relative to the image, which may bias the model towards favoring certain object-to-image size ratios at inference time.

As shown in ~\autoref{bbox_analysis_table}, images with sizes closer to 1024×1024 pixels are better understood by SAM. Since SAM only accepts images at this fixed size, the more the input image must be rescaled, the more its quality and interpretability may degrade.

Additionally, the table suggests that objects should not be extremely large or very small to be correctly annotated. Finding a balanced object-to-image size ratio appears crucial for achieving optimal performance.

\begin{table}[htbp]
    \centering
    \scalebox{0.72}{
    \begin{tabular}{l *{5}{S[table-format=1.3]}}
    \toprule
    & {500x500} & {1000x1000} & {2000x2000} & {3000x3000} & {4000x4000} \\
    \midrule
    Object: 26x26 & 0.808 & \textbf{0.828} & \textbf{0.828} & \textbf{0.828} & \textbf{0.828} \\
    Object: 52x52 & 0.927 & 0.930 & 0.930 & 0.930 & 0.930 \\
    Object: 78x78 & \textbf{0.956} & 0.944 & 0.933 & 0.933 & 0.933 \\
    Object: 104x104 & \textbf{0.953} & \textbf{0.959} & 0.929 & 0.929 & 0.929 \\
    Object: 130x130 & \textbf{0.968} & \textbf{0.958} & 0.932 & 0.920 & 0.920 \\
    Object: 156x156 & \textbf{0.967} & \textbf{0.969} & 0.927 & 0.903 & 0.903 \\
    Object: 182x182 & \textbf{0.969} & \textbf{0.974} & 0.936 & 0.912 & 0.873 \\
    Object: 208x208 & \textbf{0.969} & \textbf{0.977} & 0.929 & 0.892 & 0.871 \\
    Object: 234x234 & \textbf{0.975} & \textbf{0.970} & \textbf{0.957} & 0.921 & 0.868 \\
    Object: 260x260 & 0.611 & \textbf{0.976} & 0.946 & 0.898 & 0.879 \\
    Object: 286x286 & 0.612 & \textbf{0.978} & 0.940 & 0.911 & 0.898 \\
    Object: 312x312 & 0.542 & \textbf{0.972} & \textbf{0.954} & 0.907 & 0.888 \\
    Object: 338x338 & 0.429 & \textbf{0.976} & \textbf{0.951} & 0.922 & 0.910 \\
    Object: 364x364 & 0.566 & \textbf{0.969} & \textbf{0.965} & \textbf{0.979} & 0.899 \\
    Object: 390x390 & \textbf{0.974} & \textbf{0.979} & \textbf{0.976} & 0.919 & 0.913 \\
    Object: 416x416 & 0.000 & \textbf{0.976} & \textbf{0.974} & 0.903 & 0.891 \\
    Object: 442x442 & 0.459 & \textbf{0.983} & \textbf{0.976} & 0.931 & 0.907 \\
    \bottomrule
    \end{tabular}
    }

    \caption{Impact of object size and image resolution on SAM detection scores. The table shows detection scores for objects of varying sizes across different image resolutions (from $500 \times 500$ to $4000 \times 4000$ pixels). Bold values indicate the maximum score in a row or scores $\geq 0.95$. Results demonstrate that SAM performs optimally when objects maintain a balanced size relative to the image resolution, with sizes closer to $1024 \times 1024$ yielding higher and more consistent scores.}
    \label{bbox_analysis_table}
\end{table}

\subsection{Enabling Seamless Integration of SAM with Fiji: The Role of Appose and SAMJ}

The integration of SAM into the Fiji ecosystem via SAMJ is facilitated by Appose\footnote{\url{https://github.com/apposed/appose}}, a software framework designed to streamline inter-process communication. Within SAMJ, Appose orchestrates communication between the Java environment--where Fiji and SAMJ reside--and the Python environment, which runs SAM’s inference via PyTorch.

Appose employs process pipes to transmit variables between Java and Python and shared memory to exchange images and tensors efficiently, thus avoiding unnecessary data duplication. This architecture neatly segregates computational tasks, mitigating the risk of memory conflicts and enabling multiple models to be run concurrently without interference. Once a model is deselected, the associated processes are automatically terminated, ensuring a clean and self-contained execution environment.

In addition to managing communication, Appose automatically handles the installation and setup of required software environments. Using Micromamba, it can create and maintain isolated environments with all necessary dependencies without requiring user intervention. This fully automated process ensures a seamless user experience, simplifying the deployment and use of SAM within Fiji.

Appose creates a robust foundation for SAMJ to function as a powerful Java-to-Python bridge. This allows users to integrate SAM's advanced segmentation capabilities, enabling the annotation of images in the Fiji ecosystem, benefiting from all tools, plugins, and capabilities of Java.
Appose and SAMJ mark a significant checkpoint in the integration of novel, cutting-edge technologies developed in Python into the user-friendly world of Java.

\begin{figure}[h]
    \centering
    \includegraphics[width=1\linewidth]{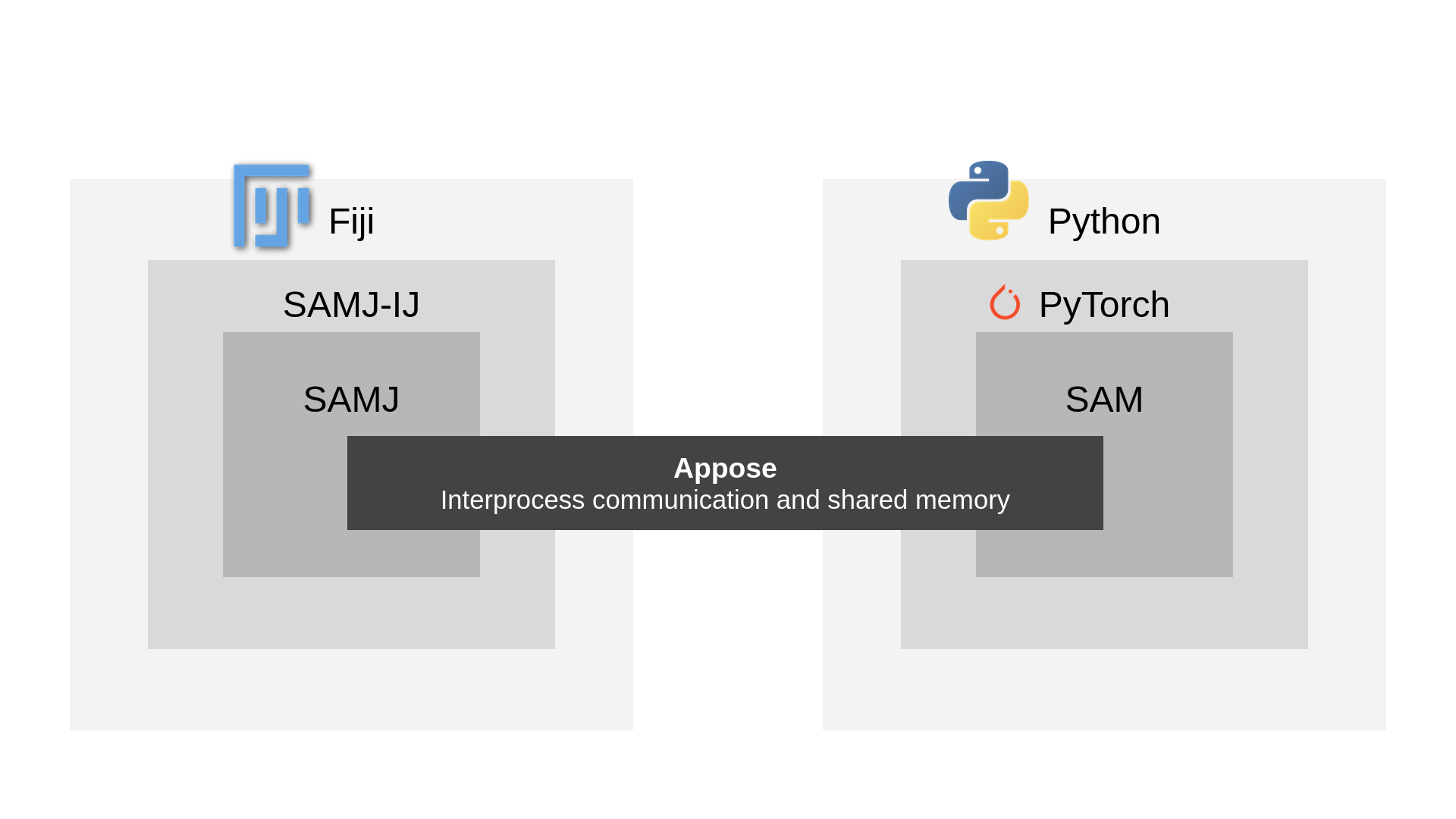}
    \caption{\textbf{Overview of the integration between SAMJ and SAM via Appose.} On the left, the Fiji/ImageJ environment hosts the SAMJ plugin, which acts as the interface for users within Java-based image processing tools. On the right, the Python with Pytorch runs the SAM model for segmentation tasks. Appose bridges these two environments, enabling seamless inter-process communication between Java and Python. It facilitates variable exchange via pipes and shared memory, ensuring efficient handling of images/tensors while avoiding memory duplication. This architecture encapsulates processes, preventing conflicts, and allows for the concurrent execution of multiple models.}
    \label{fig:appose_diagram}
\end{figure}

\subsection{Extending SAMJ for Large-Scale and Multi-dimensional Image Annotation with BigDataViewer} 
Building on the SAMJ plugin, we developed an extension for BigDataViewer, a core Fiji tool for visualizing large and multi-dimensional images \cite{pietzsch2015bigdataviewer}. This extension enables users to annotate these types of images by drawing rectangular prompts directly on the displayed image, which are immediately processed by the selected SAM model from the SAMJ Annotation control panel. The resulting polygons are rendered over the image and stored with their spatio-temporal coordinates. Polygons are displayed only when the user navigates back to the corresponding view angle and time point, but to simplify navigation, the extension memorizes view configurations, allowing users to cycle through and revisit them seamlessly. New view configurations trigger automatic image embedding, and additional features such as an undo/redo mechanism and mask export to Fiji further enhance usability.

The software is a lightweight library designed to integrate with an existing BigDataViewer instance, enabling users to add prompts directly. It allows client programs to define custom routines triggered when prompts are entered or new polygons are created. Our implementation facilitates communication between BigDataViewer and SAM models using this mechanism, while also enabling client programs to override default behaviors—for instance, converting prompts into polygons or customizing how polygons are rendered. This flexibility has been applied in Mastodon (a large-scale tracking framework for large, multi-view images), where polygons are replaced with spots for tracking, and in Labkit, where SAM-generated polygons are used for pixel classification labels.

This new library for BigDataViewer is central to enabling efficient annotation of large and multi-dimensional images. By connecting SAMJ with BigDataViewer, users can seamlessly annotate much larger datasets, including those in 3D, significantly expanding the scope of bioimage analysis.

\section{Supplementary Materials}\label{secB1}

\subsection{Links to Videos}

\begin{itemize}
  \item SAMJ: Nuclei Segmentation using Fiji’s capabilities and BatchSAMize: \url{https://www.youtube.com/watch?v=vpOzqLyxzrk}
  \item SAMJ: Tumor Area Quantification and Nuclei Analysis with SAMJ and StarDist plugins for Fiji: \url{https://www.youtube.com/watch?v=4JhSEtDxY9g}
  \item SAMJ: Accelerating Annotation of Bacterial Motility: \url{https://www.youtube.com/watch?v=10PTr5DkgBc}
  \item SAMJ: Integration with Labkit: \url{https://www.youtube.com/watch?v=2l0cyjNm80o}
\end{itemize}

\end{appendices}


\bibliography{sn-bibliography}

\end{document}